\documentclass[reprint]{JASAnew}
\usepackage[T1]{fontenc}
\usepackage[protrusion=true,expansion=true]{microtype}
\usepackage{graphicx}
\usepackage{booktabs}
\usepackage{array}
\usepackage{url}
\usepackage{amsmath}
\usepackage{placeins}
\hypersetup{
  pdftitle={Rhythm of the Deep: Two-tier acoustic organization of sperm-whale codas from click waveforms to coda sequences},
  pdfauthor={Mudit Sinha and Sanika Chavan},
  hidelinks
}

\renewcommand{\today}{16 November 2026}

\makeatletter
\def\section{\@startsection{section}{1}{\z@}{9 pt}{4 pt}{\noindent\boldmath\bigsmall\bfseries\sffamily}}
\def\subsection{\@startsection{subsection}{2}{\z@}{9 pt}{4 pt}{\noindent\boldmath\bigsmall\normalfont\bfseries\sffamily}}
\makeatother

\begin{document}
\setlength{\abovecaptionskip}{3pt}
\setlength{\dblfloatsep}{5pt}
\setlength{\textfloatsep}{4pt}
\setlength{\dbltextfloatsep}{4pt}
\setlength{\belowcaptionskip}{0pt}
\makeatletter
\let\pre@bibdata\@empty
\let\auto@bib\@empty
\makeatother

\title[Rhythm of the Deep]{Rhythm of the Deep: Two-tier acoustic organization of sperm-whale codas from click waveforms to coda sequences}
\author{Mudit Sinha}
\email{muditsinha01@gmail.com}
\affiliation{Independent Researchers, San Bruno, California 94066, United States}
\author{Sanika Chavan}
\email{sanikac10@gmail.com}
\affiliation{Independent Researchers, San Bruno, California 94066, United States}
\date{16 November 2026}

\begin{abstract}
Sperm-whale codas are usually described by the number and timing of their clicks. These measures reveal patterns in whale communication, but do not describe how individual clicks sound. We ask what click waveforms add beyond timing and how acoustic information is organized within codas and across sequences of codas. We analyze 1,952 codas, using frozen audio encoders to identify recurring forms of clicks and whole codas independently. Direct spectral and envelope measurements then test whether the click forms differ in recorded sound. Within codas, click-form composition improves the measured prediction of whole-coda form beyond timing alone. Click spacing also remains related to whole-coda sound when the same click forms are present. Neighboring click forms help predict the next click in the annotated sample, although their order does not improve the tested classification of whole codas. Across recorded sequences, the second preceding whole-coda form improves prediction beyond the immediately preceding form. The gain exceeds two broad first-order references, but stricter controls do not establish a general higher-order rule. Together, these findings show different organizing relationships at the two scales. Within codas, click sound and rhythmic spacing help distinguish complete vocalizations. Across codas, earlier whole-coda forms help predict what follows.
\end{abstract}

\maketitle

\section{Introduction}

Sperm-whale codas are short sequences of broadband clicks. They are commonly described by the number of clicks they contain and the intervals between them, known as inter-click intervals (ICIs) \citep{watkins1977sperm,rendell2003vocal}. These timing measures have revealed differences among individuals, social units, and vocal clans, as well as variation during vocal exchanges \citep{gero2016individual,oliveira2016sperm,hersh2022evidence,sharma2024contextual}. However, two codas with similar timing patterns need not sound alike. Individual clicks can also differ in their frequency content and in the shape of their amplitude envelopes \citep{goold1995time,schulz2009offaxis}. These differences are not represented by click count or ICIs, which describe when clicks occur rather than their acoustic form. Understanding how individual codas differ therefore requires considering both the sounds of their clicks and how those clicks are arranged in time.

\begin{figure*}[!t]
\centering
\includegraphics[width=0.72\textwidth]{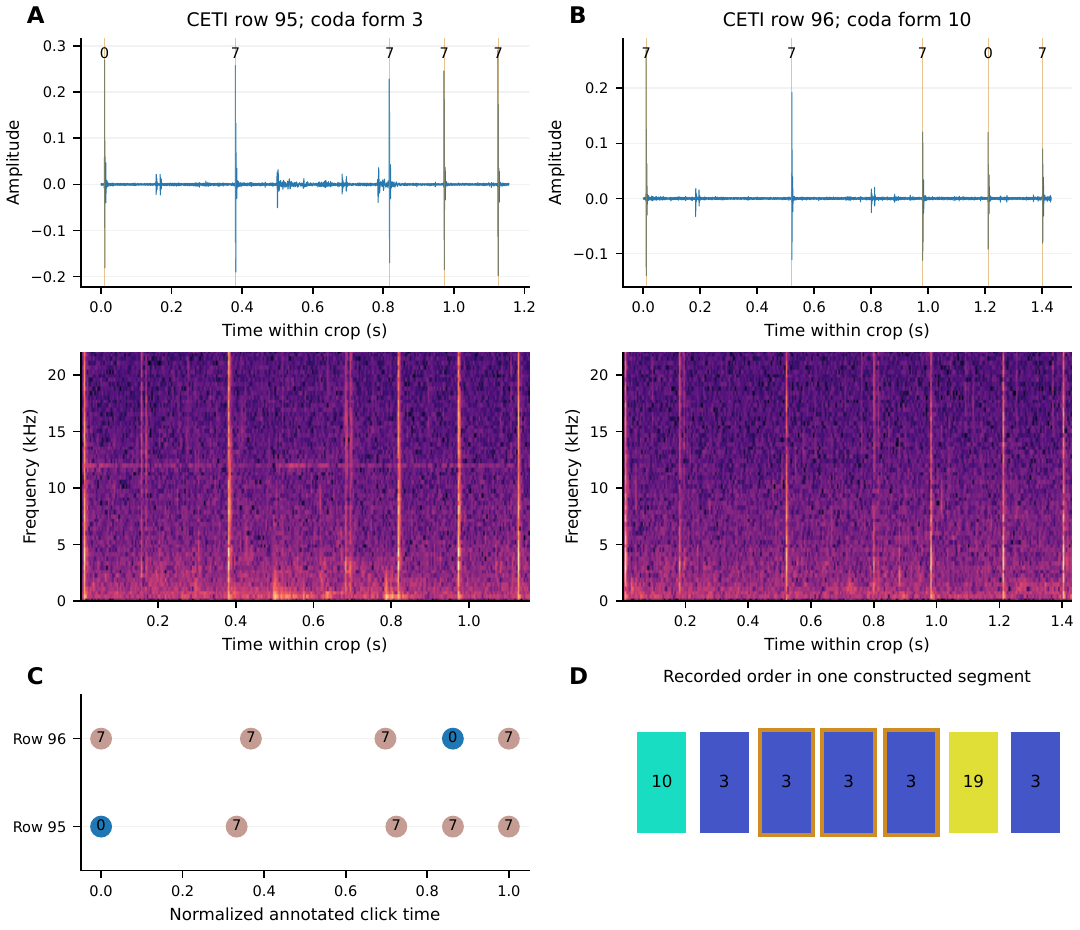}
\caption{Acoustic examples and two levels of organization. (A,B) Waveforms and spectrograms of strict-cohort codas 95 and 96, assigned to whole-coda forms 3 and 10. Lines and numbers mark annotated clicks and repaired click-form labels. (C) The codas share the multiset $[0,7,7,7,7]$, but differ in timing, click order, and other acoustic properties. This illustration does not isolate timing alone. (D) Seven consecutive whole-coda forms in a constructed sequence segment; the outlined positions illustrate a three-coda context. Forms are computational labels, not published coda types or verified perceptual categories.}

\label{fig:acoustic}
\end{figure*}

Figure~\ref{fig:acoustic} illustrates why click sound and timing must be examined together. The two codas in panels A--C contain the same five assigned click forms, yet their timing, order, and waveforms differ. Knowing which forms occur does not tell us when the clicks occur. Knowing their timing does not tell us how the clicks sound. These are different parts of a coda's acoustic description.

We first ask what the waveforms of individual clicks add beyond click count and timing. We then ask the complementary question: does click spacing still relate to whole-coda sound when the same click forms are present? Finally, we test whether the form of one click helps predict the next. These comparisons distinguish the sounds that make up a coda, the spacing between them, and the order in which their forms occur.

Individual codas also occur as parts of longer recorded sequences. Here the question changes. Instead of asking how clicks combine within one coda, we ask how complete codas relate to the codas around them. Knowing the acoustic makeup of one coda does not tell us whether earlier codas help predict the next. Previous studies have found relationships across coda exchanges \citep{sharma2024contextual,assayag2026coda}. We build on that work by studying whole-coda forms identified from their recorded sound, rather than from timing alone. We test whether knowing two preceding whole codas improves prediction of the next one over knowing only its immediate predecessor. This lets us compare what order contributes within a vocalization with what it contributes across complete vocalizations.

To answer these questions, we must compare sounds without first assigning forms from click count, timing, or sequence position. We use frozen pretrained audio encoders to identify recurring forms separately for clicks and complete codas. The analysis includes 1,952 codas and extends an earlier preprint \citep{sinha2026rhythm}. The encoders allow us to compare whole waveforms, but their assigned forms still need an acoustic check. We therefore measure frequency content and amplitude envelopes directly from the recordings. We also compare forms across held-out recordings and matched conditions. Because click forms and whole-coda forms are identified independently, any relationship between them must be measured rather than built into the labels.

Our findings follow these questions. Within codas, click-form composition improves the measured description beyond timing alone, and spacing remains related to whole-coda sound when composition is matched. Neighboring click forms also help predict one another in the annotation-based test. Yet adding their order does not improve the tested classification of the complete coda. Across recorded sequences, the second preceding whole-coda form improves prediction beyond the immediate predecessor. Thus, order does not play one identical role at both scales. Within codas, we can separate the sounds and spacing of the constituent clicks from the prediction of the next click. Across codas, the relevant history consists of complete vocalizations. Together, the analyses show what these different measurements reveal about how coda acoustics are organized.

\section{Related work}

Research on sperm-whale codas has traditionally focused on click count and inter-click intervals, revealing differences among individuals, social units, and vocal clans \citep{watkins1977sperm,rendell2003vocal,gero2016individual,oliveira2016sperm,hersh2022evidence}. These measures describe when clicks occur, but leave much of their sound unrepresented. Earlier work characterized spectral and temporal properties of sperm-whale clicks, including effects of recording angle \citep{goold1995time,schulz2009offaxis}. \citet{begus2025vowels} identified vowel- and diphthong-like spectral patterns in codas. Subsequent work examined how those patterns relate to coda type and duration \citep{begus2026phonology}. These studies show that clicks differ in sound as well as timing. We ask whether recurring waveform differences help distinguish whole codas when their timing is already known. We then test whether spacing still matters when the same click forms occur, and whether one click form helps predict the next. We also measure spectral and envelope properties directly to check the learned click forms.

Codas also occur in sequences, and previous work has examined relationships among successive vocalizations. \citet{sharma2024contextual} described rhythm, tempo, rubato, and ornamentation, including variation associated with neighboring codas. \citet{assayag2026coda} showed that preceding timing features help predict later vocalizations in selected exchanges between whales. We instead study whole-coda forms identified from their recorded sound. We ask whether knowing the second preceding coda helps predict the next one after the immediate predecessor is known. This tests relationships among complete codas without using only their timing descriptions.

We need to identify recurring click and whole-coda sounds without first labeling them by timing or their position in a sequence. Automated methods and learned audio embeddings have been used to group animal vocalizations \citep{deecke2006automated,morfi2021deep,goffinet2021low,best2023deep}. Neural networks have also been used to detect sperm-whale clicks and classify codas \citep{bermant2019deep,gubnitsky2025automatic}. Pretrained audio encoders offer another way to compare recorded sounds \citep{hagiwara2023aves,ghani2023global,miron2026avex}. In sperm whales, WhAM has used learned representations for coda generation and for classifying rhythm, social-unit, and spectral categories \citep{paradise2025wham}. Here we use frozen encoders to identify click and whole-coda forms separately, then check them against direct waveform measurements. We can therefore test how click sound and timing relate within codas, and how whole codas follow one another.

\section{Methods}

We begin with the sounds of individual clicks, then examine the codas they form, and finally study the order of complete codas. First, we identify recurring acoustic forms separately for clicks and whole codas using frozen audio encoders. We test whether click sound adds to conventional timing and whether spacing remains informative when click composition is held fixed. We then examine click-form order in two ways: prediction of the next click and description of the complete coda. At the second scale, we test whether earlier whole-coda forms help predict the one that follows. Each step answers a different question. Direct waveform measurements check the acoustic meaning of the assigned forms, while held-out tests and matched controls examine alternative explanations. Table~\ref{tab:methods_overview} summarizes the comparisons.

\begin{table*}[!t]
\centering
\caption{Analyses used to examine acoustic organization within codas and prediction across codas. Each comparison uses an independent acoustic description or a control designed to test a specific alternative explanation.}
\label{tab:methods_overview}
\footnotesize
\setlength{\tabcolsep}{3pt}
\newcommand{\methodcell}[2]{\parbox[t]{#1\textwidth}{\raggedright #2}}
\begin{tabular}{@{}llll@{}}
\toprule
Analysis & Purpose & Measurement & Main control \\
\midrule
\methodcell{0.16}{Acoustic-form clustering} &
\methodcell{0.20}{Identify recurring click and whole-coda sounds} &
\methodcell{0.24}{Frozen-encoder embeddings and consensus grouping} &
\methodcell{0.25}{Multiple encoders and grouping resolutions} \\[3pt]
\methodcell{0.16}{Click-to-coda comparison} &
\methodcell{0.20}{Separate click sound from conventional timing} &
\methodcell{0.24}{Held-out association, prediction, and composition matching} &
\methodcell{0.25}{Matched shuffles and detector sensitivity} \\[3pt]
\methodcell{0.16}{Click-form ordering} &
\methodcell{0.20}{Test sequence information beyond click composition} &
\methodcell{0.24}{Bigram gain, ordered versus bag features, and position} &
\methodcell{0.25}{Within-coda shuffles and acoustic-null inputs} \\[3pt]
\methodcell{0.16}{Direct acoustic validation} &
\methodcell{0.20}{Check whether learned forms differ in measured sound} &
\methodcell{0.24}{Spectral and envelope descriptors at both levels} &
\methodcell{0.25}{Matched recording context and timing-only references} \\[3pt]
\methodcell{0.16}{Whole-coda prediction} &
\methodcell{0.20}{Test information from preceding codas} &
\methodcell{0.24}{Held-out prediction from one, two, and three previous codas} &
\methodcell{0.25}{Source exclusion, first-order and run-preserving references} \\
\bottomrule
\end{tabular}
\end{table*}

\subsection{Recordings and click extraction}

We begin with recordings and click annotations from the publicly shared sperm-whale audio collection linked by Gubnitsky and Diamant \citep{gubnitsky2024presence}. The analysis uses 33 remote WAV recordings and 30 tag FLAC recordings, whose audio was decoded losslessly for analysis. Together, these 63 recordings provide 1,952 coda crops and 10,755 annotated clicks. The supplementary recording manifest identifies each source file and its annotation sheet. The annotations supply click positions and inter-click intervals (ICIs), but not reliable date, social-unit, caller, or published coda-type labels. Relative onset times and gaps establish order within recordings. They do not establish whether consecutive codas belong to a behavioral exchange.

To study click sound, we need waveform windows around individual clicks. We extract each window from 2 ms before to 18 ms after its annotated position. Some coda crops share portions of the same native audio or repeat the same waveform. The strict cohort excludes this shared coverage and these duplicates, leaving 1,205 codas. We then remove 73 codas with overlapping click windows, leaving 1,132 codas containing 6,039 clicks. This prevents the main click comparisons from treating shared audio or overlapping windows as independent examples. The annotated times locate the clicks; they do not determine which clicks sound alike. Whole-coda sequence tests use the full corpus, subject to the amount of preceding history available.

The main click analysis uses annotated positions, so its results could depend on which events those annotations include. We therefore repeat click location with an independent envelope-based detector. We compare the event streams and separate the effects of adding or removing events from those of moving the extraction windows. This tests how click selection affects the order results. The settings and comparisons are reported in the supplement.

\subsection{Frozen-encoder representations and consensus clustering}

We identify recurring forms among click and whole-coda waveforms using pretrained audio encoders, without defining those forms from coda types or annotated click intervals. An encoder converts a waveform into a numerical representation, called an embedding, that captures acoustic features learned during its original training. Similar embeddings indicate similar sounds under that representation. The encoder parameters remain unchanged.

Different encoders emphasize different acoustic properties, so we use multiple representations. We use eight pretrained encoder families: AVES \citep{hagiwara2023aves}, AVEX \citep{miron2026avex}, HuBERT \citep{hsu2021hubert}, BEATs/OpenBEATs \citep{chen2023beats,bharadwaj2025openbeats}, wav2vec 2.0 \citep{baevski2020wav2vec}, Whisper \citep{radford2023whisper}, Perch \citep{ghani2023global}, and VampNet \citep{garcia2023vampnet}. Click windows and complete-coda waveforms are encoded separately, so neither set of forms is constructed from the other.

Embeddings describe similarity but do not identify discrete forms. We standardize each encoder's features and reduce them using principal component analysis (PCA) \citep{jolliffe2002pca}. The primary click and coda fits use at most 32 and 64 components, respectively. We then apply K-means clustering \citep{lloyd1982least}, which groups recordings by their proximity to learned centers. The number of centers determines how finely we divide the recordings; it does not assume a particular number of biological categories.

We combine encoder-specific groupings through consensus clustering \citep{strehl2002cluster,monti2003consensus,fred2005combining}. For each pair of recordings, we measure how often encoder views assign them to the same group, then apply average-linkage clustering to those agreements. This identifies similarities shared across representations.

The consensus assignments initially changed when the same waveforms were stored in a different row order. To make the result reproducible, we sort waveforms using a fixed SHA-256 hash of each event identity before clustering. We restore their original identities before any analysis of recorded order. Standardization, PCA, and K-means also use the fixed training-row order. After this repair, permuting storage order gives an adjusted Rand index (ARI) of 1.0 for both click and whole-coda forms. This checks reproducibility of the assigned groups, not whether they are biological categories.

The main analysis identifies 12 click forms from 20 curated encoder views and 32 whole-coda forms from 23 views. The storage-order repair preserves the original partitions within each view. For fully source-held-out tests, we fit 18 click views and 21 whole-coda views using training sources only. The supplement reports the encoder settings and tests with other grouping resolutions.

\subsection{Click-to-coda association and timing comparisons}

The first comparison asks whether the sounds of individual clicks tell us anything about the complete codas that contain them. Because click and whole-coda forms are learned separately, an association between them is not automatic. For each click, we pair its assigned form with the independently assigned form of its parent coda. We measure their association using normalized mutual information (NMI) \citep{strehl2002cluster}. This tells us whether some click forms occur more often in some whole-coda forms than in others. It does not yet tell us whether click composition improves prediction beyond conventional timing.

We test this association on recordings that were not used to identify the forms. In four whole-source-held-out folds, the training sources define standardization, PCA, clustering, and consensus using 18 click and 21 coda representation views \citep{roberts2017cross}. The held-out clicks and codas then receive assignments from those fitted procedures. For a reference comparison, we permute the parent-coda labels among intact codas matched by recording and click count. This breaks the observed link between click form and parent-coda form without separating clicks that belong to the same coda. Treating those clicks as independent rows would give a different, less appropriate reference.

The NMI test asks whether the two sets of labels are related. Our next test asks a more direct question: do click waveforms add anything when timing is already known? We represent each coda by the proportions of its 12 click forms. This captures which click sounds are present without using their order. We compare three predictors of whole-coda form: conventional timing features alone, click-form composition alone, and both together. We score each predictor on the same four held-out sources. We also resample and test at the source level, because many codas from one recording source do not amount to many independent sources of evidence. Comparing the combined predictor with timing alone measures the added value of click composition in the available data.

The reverse question matters as well. If two codas contain the same click forms in the same quantities, they can still differ in the spacing of those clicks. We therefore compare codas with identical click-form multisets. For each pair, we measure differences in normalized ICI patterns and differences between their whole-coda acoustic embeddings. We repeat this comparison within the same recording and retain all available ICI slots. An association would show that matching the click sounds does not remove the relationship between spacing and whole-coda acoustics. The codas may still differ in other sound properties, so this is a test of association rather than a timing-only intervention.

\subsection{Click-form ordering and acoustic controls}

Knowing which click forms occur in a coda does not determine their order. Two codas can contain the same forms, in the same quantities, but place them differently. We first ask whether the form of one click helps predict the next after we account for which forms are present. We then ask whether that ordered information improves the description of the complete coda. These are related questions, but a positive answer to one does not require a positive answer to the other.

We fit a bigram predictor, which uses the preceding click form to predict the next one. A context-free predictor provides a comparison that does not use the preceding form. Both models are evaluated on the same actual next-click targets in held-out sources. We calculate their base-2 log probabilities and normalize once by the number of targets, giving a gain in bits per click. To ask whether the gain depends on recorded order, we shuffle the click-form labels within each held-out coda 999 times. Every shuffle keeps the coda length and the counts of each form unchanged, but changes which forms follow one another. We compare the recorded gain with this reference and apply Holm correction to the four source-fold tests.

A bigram gain would show that neighboring click forms are related. It would not show that their order helps identify the acoustic form of the complete coda. We therefore make a separate comparison. One set of features counts how often each click form occurs. A second set also records click positions and adjacent form pairs. We compare them in whole-coda classification and continuous acoustic reconstruction. We also test whether particular click forms tend to occupy particular positions. Together, these measurements distinguish a local next-click relationship from an ordering pattern that improves the description of the whole vocalization.

Even a gain over label shuffles need not arise from a relationship specific to the recorded click waveforms. Properties such as spectral shape or amplitude envelope could produce a similar gain under a controlled replacement. We therefore test spectrum-matched noise, envelope-shaped noise, cross-coda click replacement, and waveform-order shuffling. The controls change the waveform information in different ways while retaining selected properties of the original sounds. For each applicable input, we identify new acoustic forms and repeat the prediction test on the same strict cohort and common 18-view panel. This asks whether the recorded next-click gain exceeds the gains available from the tested altered sounds, rather than treating any nonzero gain as sufficient evidence.

For the real gain to pass our acoustic-specificity screen, it must exceed the largest control gain by at least 0.002 bits or 15\% of the larger absolute gain. It must also exceed that gain by a factor of 1.10. This is an effect-size screen, not a significance test. The highest-rate Perch and VampNet destructive controls are unavailable. The supplement gives further controls, including persistence tests and independent click detection.

\subsection{Direct acoustic measurements and recording-context controls}

The encoders group waveforms by learned acoustic similarity, but their labels do not tell us what differs in the sound. We therefore return to the recordings and measure each click's frequency content and amplitude envelope. These measurements are not used to make the groups. Instead, they let us test whether the learned click forms also differ in familiar acoustic properties. We examine individual clicks first, then ask whether summaries of those measurements help identify the forms of complete codas. A waveform can also change with source orientation, propagation, or recording equipment. We therefore compare clicks under matched recording conditions, so differences between recordings are not mistaken for differences between click forms.

\paragraph{Click-level spectral and envelope measurements}

We begin with the 6,039 clicks in the strict cohort. For each click, we measure eight spectral properties. These are peak frequency, spectral centroid, bandwidth, 85\% spectral rolloff, spectral slope, flatness, entropy, and the ratio of high- to low-frequency energy \citep{peeters2011timbre}. We also measure four properties of its amplitude envelope: width, rise time, fall time, and decay time. Together, these 12 descriptors capture spectral shape and how click amplitude changes over time.

We examine RMS level, signal-to-noise ratio, and crest factor separately because these measurements can be strongly affected by how a click was recorded. All descriptors are computed at each recording's native sample rate, here 60 or 96 kHz. We first compare how well the spectral and envelope shape measurements, the level/noise measurements, and both together distinguish the click forms on held-out sources. We then normalize shape measurements within codas and within recording-and-native-rate groups. These comparisons ask whether the differences between forms remain when we focus on clicks recorded under more closely matched conditions, rather than on differences between recordings.

The direct measurements are evaluated against a matched reference. We shuffle click-form labels within groups defined by recording and native sample rate, then repeat the prediction test. The shuffle preserves the distribution of recording conditions in each group, but removes the observed correspondence between a click's measurements and its assigned form. We compare balanced accuracy for the real and shuffled labels. We also ask which individual shape descriptors differ among forms, correcting those comparisons with the Benjamini--Hochberg false-discovery-rate procedure \citep{benjamini1995controlling}. Throughout this step, the original encoder-assigned labels stay fixed; the direct acoustic measurements are checks on those labels, not inputs used to create them.

\paragraph{Whole-coda acoustic summaries and timing references}

Differences among individual clicks do not automatically mean that a few waveform measurements can identify the form of a complete coda. A coda contains several clicks, and its overall sound may depend on their combination as well as their spacing. To test what the direct click measurements capture at this larger scale, we summarize all twelve shape descriptors within each coda. For each descriptor, the median measures the typical click property, the interquartile range measures how much it varies, and the trend describes how it changes across click positions. These give 36 summaries of click sound within a coda. We compare them with click count, ICIs, and duration, which describe conventional timing. The comparison tests whether simple measured waveform properties add to timing in identifying the independently assigned whole-coda forms.

We test three predictors of whole-coda form: timing alone, waveform summaries alone, and their combination. We use source-held-out balanced accuracy so that each form contributes equally to the score \citep{brodersen2010balanced}. As a reference, we shuffle whole-coda form labels within groups matched by recording, native rate, and click count. We also test whether individual descriptors differ among the forms, with correction for multiple comparisons. These tests separate two questions: whether whole-coda forms show measured acoustic differences, and whether adding these summaries improves classification over timing. The current ledger contains no published coda-type labels, so we do not infer agreement with or subdivision of those types.

\subsection{Whole-coda sequence prediction and first-order controls}

The preceding tests describe the clicks within a coda and the acoustic forms of complete codas. They do not tell us how those complete forms appear in recorded sequences. We now change the unit of analysis from individual clicks to whole codas. Specifically, we ask whether knowing the form two codas earlier improves prediction of the next form after the immediately preceding coda is already known. The whole-coda forms were identified without using their recorded order, so the sequence relationship is not built into their labels.

A sequence test first needs an order in which codas can be compared. The 63 recordings provide relative onset times and gaps, but not verified behavioral exchanges. We group nearby codas using an eight-second inter-coda gap threshold, producing 182 operational groups. Ambiguous chronology is split further, giving 195 sequence segments. Within each segment, we retain the observed order without joining uncertain gaps. The boundaries are operational rather than verified natural bouts. We also define four conservative source pools for broader exclusions. These are not identified callers or social units. Keeping those distinctions clear lets us measure prediction from recorded order without assigning a behavioral interpretation to the segments.

We measure how much each additional preceding coda helps by fitting four predictors. Order 0 ignores sequence history. Order 1 uses the immediately preceding form, order 2 uses the preceding two forms, and order 3 uses three. We score all four on the same target positions so that a change in score is not caused by evaluating different codas. The common set has three previous codas available and contains 1,478 target positions in 114 segments. When a history is too rare in the training data, the predictor falls back to a shorter history \citep{katz1987estimation}. For each target we calculate the negative base-2 log probability assigned to the observed form. Lower loss means better prediction. The difference between two model orders gives the added information from the extra preceding coda, measured in bits per coda.

The main acoustic-form inventory is fitted to the full corpus, although its clustering uses no chronological order. That choice makes the main test a measure of prediction within the available collection. To ask how well the relationship transfers, we repeat the prediction comparisons while holding out entire recordings and then whole sources. We also perform a stricter source-held-out test in which standardization, PCA, per-view K-means, choice of $K$, and consensus are fitted without the test source. Its codas are assigned forms only after this fit. These comparisons separate a gain measured within the corpus from a gain that would also hold for an entirely unseen source.

\paragraph*{First-order sequence controls.}

An improvement from the second preceding coda does not, by itself, show a higher-order sequence rule. Different segments may have different form frequencies or different patterns between adjacent codas. Pooling these segments can make a second predecessor informative even when the relationships within each segment involve only neighboring codas.

We measure the information provided by the second preceding form once the immediately preceding form is already known. This is conditional mutual information:
\begin{equation}
I(C_t;C_{t-2}\mid C_{t-1})
=H(C_t\mid C_{t-1})-H(C_t\mid C_{t-1},C_{t-2}),
\label{eq:coda_conditional_information}
\end{equation}
Here $C_t$ is the current whole-coda form and entropy $H$ measures uncertainty in bits \citep{cover2006elements}. The difference asks how much uncertainty is reduced by adding the second preceding form after the first is known. We estimate it using the Nemenman--Shafee--Bialek estimator \citep{nemenman2002entropy}. We then compare the result with three first-order references. The corpus-global reference uses shared neighboring-coda transitions, while the hierarchical reference allows differences among segments. The strict segment-local reference more closely preserves the observed neighboring-coda transitions within each segment. Each reference tests a different way that local transition patterns might produce an apparent benefit from a second predecessor. The supplement also reports a separate pooled first-order calibration of the information estimator. It uses a different fitted reference from the corpus-global comparison here, so its outcome must be read separately.

A further possibility is that repeated runs of the same form account for part of the prediction gain. We therefore perform a separate sensitivity test that preserves each segment's form counts and the lengths of its repeated-form runs. This asks how much additional order-2 prediction remains when local persistence is retained. It is not a general first-order reference, so we interpret it separately from the three sequence controls. The run analysis uses 1,605 order-2 target positions in 127 segments, rather than the 1,478 common-history positions used in the primary comparison. The supplement gives the reference constructions and uncertainty estimates.

\section{Results}

\subsection{Click acoustics and timing provide complementary information}

\begin{figure*}[!t]
\centering
\includegraphics[width=0.77\textwidth]{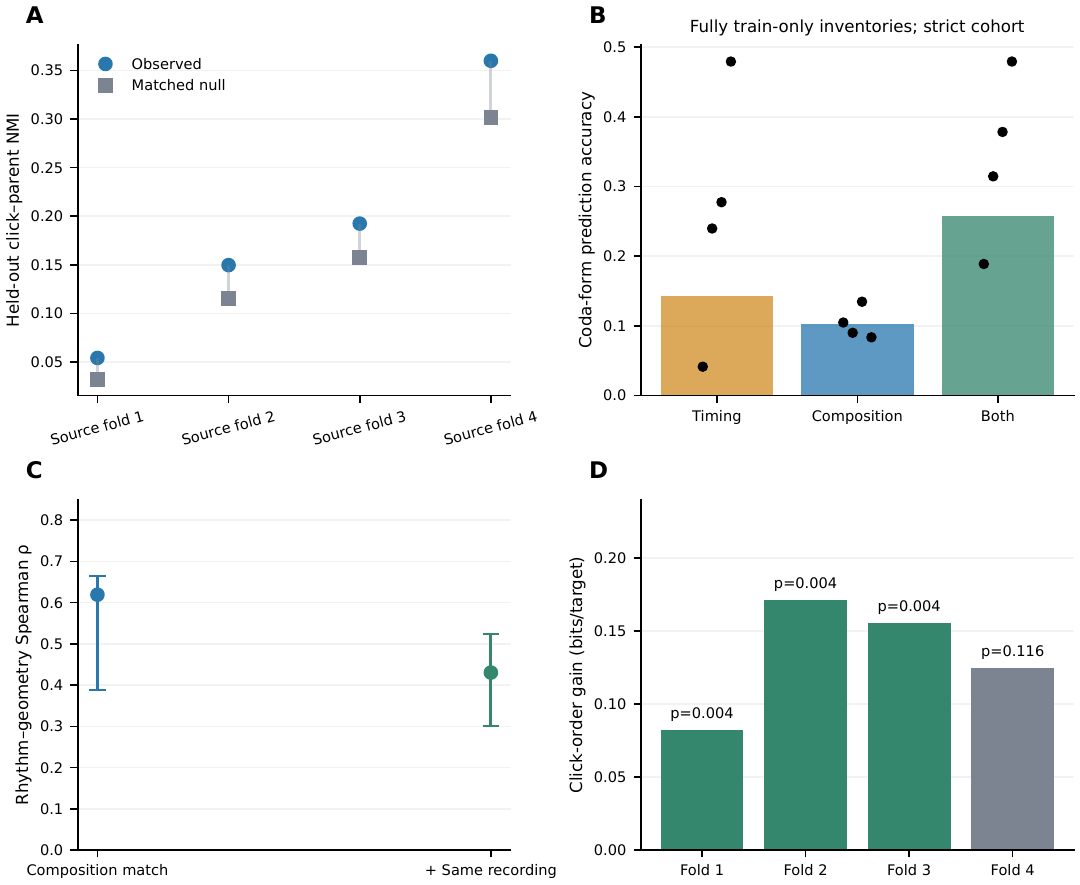}
\caption{Strict lower-tier comparisons. (A) Held-out click--parent-coda NMI and an intact-coda label-permutation reference within recording and click count. (B) Prediction accuracies from timing, click composition, and both. The combined-minus-timing gain is 0.1148, with exact source-swap $p=0.125$. (C) ICI-pattern and whole-coda embedding distances among codas with identical click-form multisets, including recording-matched pairs. (D) Corrected next-click prediction against within-coda order permutations in the 20-view annotation analysis. Positive gain does not by itself establish a biological sequencing rule.}
\label{fig:lower}
\end{figure*}

Click waveforms add information that click count and timing do not describe. We first test this by predicting whole-coda form from timing, click-form composition, and both. For 1,132 codas, timing alone gives held-out accuracy of 0.1431, and click-form composition alone gives 0.1025. Using both raises accuracy to 0.2580 (Fig.~\ref{fig:lower}B). The gain over timing alone is 0.1148, with a whole-source bootstrap interval of $[0.0523,0.1437]$. The exact four-source sign-swap test gives $p=0.125$, so the gain is not established as a source-level effect. Composition alone is weaker than timing, but combining the two gives the best prediction in the measured sample. The result concerns the sounds of the clicks present in a coda, not their order.

A second test asks whether the forms of individual clicks are associated with the forms of the codas that contain them. Unlike the composition predictor, this test compares the two sets of independently assigned labels directly. We measure their association using normalized mutual information (NMI) after identifying both sets of forms without the test source. NMI ranges from 0.0542 to 0.3598 across the four source folds (Fig.~\ref{fig:lower}A). Permuting parent-coda labels among intact codas matched by recording and click count gives reference values from 0.0320 to 0.3016. The median difference is 0.03453, and all four conditional tests remain positive after Holm correction. The permutation keeps clicks from the same coda together rather than treating them as independent. Thus, the link between click sound and whole-coda form is also visible when the forms are identified separately on training sources.

Click sound is not the whole description. We next hold click composition fixed and ask whether spacing remains related to how the complete coda sounds. We compare 1,118 coda pairs matched by click composition and recording. Pairs with larger differences in inter-click intervals also differ more in their whole-coda acoustic representations. The association is $\rho=0.4304$ (composition-group bootstrap 95\% CI $[0.3020,0.5230]$; within-group permutation $p=0.001996$; Fig.~\ref{fig:lower}C). The pairs can differ in other acoustic properties as well. Even so, the result shows that matching click forms does not remove the relationship between spacing and whole-coda sound. The two comparisons therefore point to complementary contributions from click waveform composition and rhythmic spacing.

\subsection{Click and whole-coda forms have measurable acoustic differences}

\begin{figure*}[!t]
\centering
\includegraphics[width=0.75\textwidth]{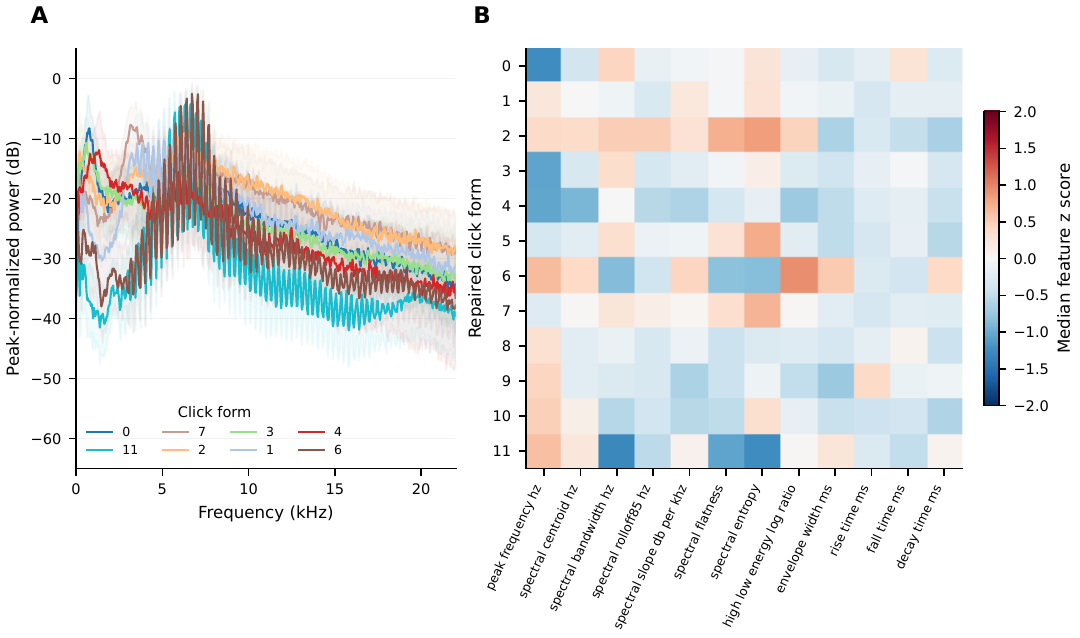}
\caption{Direct acoustic measurements of the repaired click forms. (A) Peak-normalized spectra for the eight most frequent forms over the common 0--22-kHz band. Curves show medians and shaded interquartile ranges. (B) Standardized medians of twelve spectral and envelope descriptors across all twelve forms. Inference uses source-held-out prediction and recording/native-rate matched controls.}
\label{fig:physical}
\end{figure*}

The recurring click forms differ in their recorded waveforms, not just in the encoder representations. This matters because the form labels were learned from embeddings; they were not defined from the physical descriptors tested here. The frequency content and amplitude envelopes show measurable differences (Fig.~\ref{fig:physical}). All twelve descriptor omnibus tests pass false-discovery-rate correction. Among the most frequent forms, 54 of 336 supported pair--feature comparisons also pass correction. The measurements show which aspects of the recorded sound distinguish the learned click forms.

The differences remain informative when we compare clicks under matched recording conditions. Direct waveform measurements predict click-form identity with balanced accuracy of 0.1380, compared with 0.1004 after shuffling labels within recording and native-sample-rate groups ($p=0.009901$). In a separate source-held-out test, level and noise measurements reach 0.1310, spectral and envelope shape measurements 0.2031, and both together 0.2547. Normalizing shape measurements within each coda reduces accuracy to 0.1108 (Supplementary Fig.~S1). The forms therefore contain measurable spectral and envelope information, while their separation also depends in part on recording context. This is why the matched tests matter alongside the raw descriptor differences.

The same direct measurements tell a more limited story for complete codas. Timing alone gives balanced accuracy of 0.0909, the 36 waveform summaries give 0.0387, and their combination gives 0.0860. All three exceed their matched shuffled means, but adding the waveform summaries to timing changes accuracy by only $-0.0049$ (source-bootstrap CI $[-0.0623,0.0187]$). Nineteen of 36 descriptor omnibus tests pass correction, while none of the 1,512 eligible pair--feature comparisons passes global correction. The direct measurements therefore distinguish individual click forms more clearly than they identify whole-coda forms. The 36 coda summaries retain typical click properties, their variation, and simple trends, but not the full detail of the recorded waveform. Their limited prediction gain does not mean that complete codas lack acoustic differences. It means that these particular summaries do not improve whole-coda classification beyond timing in the tested model. This is why we examine the clicks and complete codas separately.

\subsection{Click and whole-coda forms show different ordering relationships}

We have shown what click sound and spacing contribute to individual codas. We now ask what the recorded order of acoustic forms adds. There are two separate questions within a coda: can one click form predict the next, and does the sequence of click forms help identify the whole vocalization? The comparison across codas asks something else: can earlier complete codas predict the form of the coda that follows? These tests allow us to compare the role of order without assuming it must work the same way at both scales.

Within the annotated codas, knowing one click form helps predict the next. The corrected gain over within-coda order shuffles is 0.1334 bits per actual target (Fig.~\ref{fig:lower}D). The four source gains are 0.0822, 0.1716, 0.1554, and 0.1246 bits; three of the four source-fold tests pass Holm correction. The gain does not come simply from which forms are present, since the shuffles preserve their counts. In a separate matched 18-view acoustic-control panel, real click order gives a gain of 0.08894 bits, compared with at most 0.01266 for the tested destructive controls. The real result passes the acoustic-specificity effect-size screen. Independent click detection does not confirm this gain, and the additional detector checks cover only one encoder family (supplement). The result therefore describes a local relationship in the annotation-defined click sequence, not an established rule for all detected clicks.

The positive next-click result does not carry over to the other within-coda question. With the repaired forms, click counts predict whole-coda form with accuracy 0.09894; adding position and bigram features lowers accuracy to 0.08392. With timing included, the accuracies are 0.14311 and 0.13074. Click form is also not reliably associated with its absolute position (positional information lift 0.000192 bits, $p=0.3234$). Thus, neighboring forms help predict the next click, but their order does not improve the tested description of the complete coda. This separates a local sequence relationship from the acoustic composition and spacing that help distinguish whole codas.

\begin{figure*}[!t]
\centering
\includegraphics[width=0.75\textwidth]{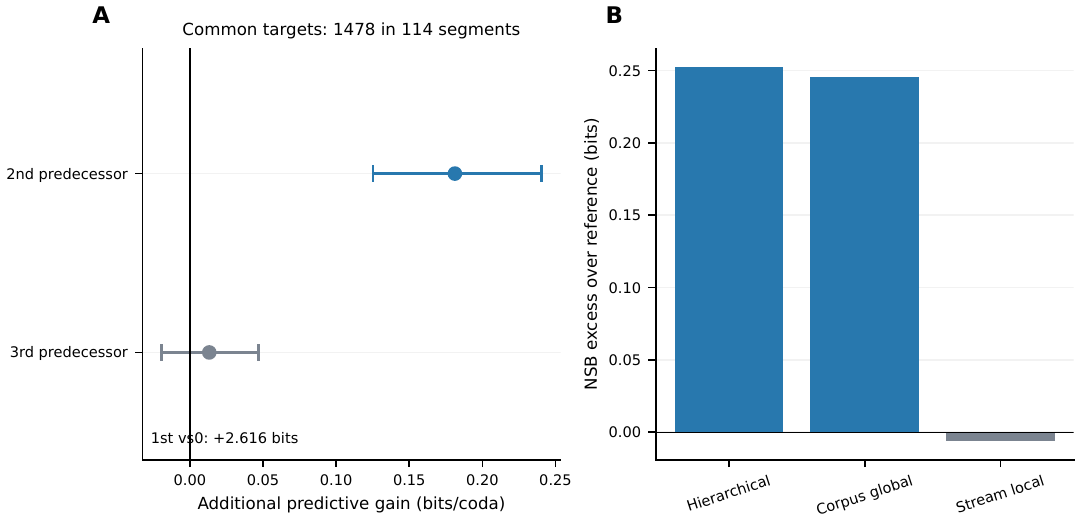}
\caption{Whole-coda sequence prediction and first-order references. (A) Added held-out prediction from a second and third predecessor on the same 1,478 targets in 114 segments. The first-predecessor gain is annotated separately. Intervals resample segments. (B) The NSB statistic relative to hierarchical, corpus-global, and strict segment-local first-order references. The strict local reference does not establish dependence beyond each segment's adjacent-coda transitions.}
\label{fig:sequencecore}
\end{figure*}

When we move from clicks to complete codas, earlier whole-coda forms also help predict what follows. Knowing two preceding forms predicts the next form better than knowing only one. We compare 1,478 target positions from 114 constructed segments. The immediate predecessor improves prediction by 2.6158 bits/coda. The second preceding form adds 0.1812 bits/coda (segment-bootstrap 95\% CI $[0.1250,0.2405]$, $p=0.000200$); 89 of 114 segments show a positive gain (Fig.~\ref{fig:sequencecore}A). A third predecessor adds only 0.0132 bits/coda, with an interval crossing zero ($p=0.2144$). The strongest added gain in the tested history lengths therefore comes from the second preceding coda. Here the relevant units are whole vocalizations, not the constituent clicks within one coda.

The size of this gain depends on which recordings and sources are held out. With the full-corpus acoustic forms fixed, it is 0.1362 bits/coda under recording holdout and 0.06834 under whole-source holdout, where the source-level uncertainty includes zero. When all acoustic-form fitting also excludes the test source, the mean gain is 0.2289 bits/coda. Its source-bootstrap interval $[-0.0156,0.3552]$ includes zero, and its exact source-swap test gives $p=0.1875$. The within-dataset gain is clear, but the source-held-out tests do not establish generalization across independent sources.

A second predecessor can help prediction even if each segment mainly follows relationships between adjacent codas. We therefore compare the measured sequence information with first-order references (Fig.~\ref{fig:sequencecore}B). The observed NSB statistic is higher than the hierarchical and corpus-global references by 0.2524 and 0.2453 bits ($p=0.001996$ for each). It does not exceed the strict segment-local reference (lift $-0.00609$ bits, $p=0.5908$). The two broad references show an excess, but the strict local reference leaves an explanation based on each segment's neighboring-coda transitions. A separate pooled first-order calibration in the supplement also shows no excess for the observed NSB statistic (upper-tail $p=1.0$). This calibration uses a different fitted reference from the corpus-global comparison above. The second-predecessor gain remains a prediction result; these controls do not establish a general higher-order sequence rule.

A separate control preserves runs of repeated forms. It leaves smaller order-2 gains of 0.0271 bits with segment holdout, 0.0231 with recording holdout, and 0.0016 with whole-source holdout, where the remaining gain is not significant. This control uses a different cohort of 1,605 targets (Supplementary Fig.~S6). Thus, earlier whole-coda forms improve prediction in the recorded sequences, while the stricter controls do not establish a general higher-order rule.

Together, the ordering tests show why it matters which acoustic unit is being predicted. Within a coda, neighboring click forms help predict each other, but including their order does not improve the tested classification of the complete vocalization. Across codas, the forms of earlier whole codas help predict the form that follows. The results do not imply that only one scale has order. They show that the measured contribution of order depends on whether we are studying the clicks that make up a coda or the complete codas in a recorded sequence.

\FloatBarrier

\section{Discussion}

Click count and timing describe when clicks occur, but not the full sounds of those clicks. Our results show why that missing information matters. Click waveform composition and rhythmic spacing contribute complementary information about individual codas. The ordering tests then show something different: the information available from neighboring click forms is not the same as the information available from earlier complete codas. The finding is not that codas have two scales. It is that studying click sound, spacing, and recorded order reveals different relationships at those scales.

\subsection{Click acoustics and timing describe different properties of codas}

Conventional descriptions tell us how many clicks a coda contains and when they occur. They do not describe the waveforms of the clicks themselves. The combined predictor shows what this leaves out: in the measured sample, adding click-form composition improves whole-coda prediction over timing alone. The matched-composition test addresses the reverse question. Even when the same click forms are present, differences in spacing remain associated with differences in whole-coda sound. The two tests therefore examine complementary properties rather than two competing ways to label the same coda.

The distinction is visible in individual recordings. Two codas may contain the same click forms but place them at different times. Two others may have similar timing while their clicks differ in sound. The first pair shows why composition is not a description of spacing; the second shows why timing is not a description of waveform sound. Both properties are needed to characterize the differences tested here. This also explains why treating click sound as merely another timing feature would miss part of the acoustic variation.

Click-form order adds a third, more local relationship. Knowing one click form helps predict the next in the annotated sample, under the tested shuffles and acoustic controls. Yet putting position and adjacent-pair information into the whole-coda classifier does not improve its accuracy. These results are compatible because they measure different things. One asks whether neighboring clicks are statistically related. The other asks whether those relationships help distinguish the entire vocalization. Within-coda organization therefore cannot be described by a single measure of click order.

\subsection{Frozen encoders identify measurable acoustic variation}

The learned click forms correspond to measurable differences in the recorded waveforms. Their spectra and amplitude envelopes differ, including under comparisons that match recording conditions. This provides an acoustic interpretation of the groups rather than relying on their separation in an encoder's representation. The same measurements are less useful when summarized across a complete coda. Their medians, variation, and trends do not improve whole-coda classification over timing in the tested model. These summaries leave out details of the full waveform and how the clicks combine. The result does not weaken the measured differences among individual clicks. It shows why identifying the sound of a click and describing the sound of a whole coda are different acoustic tasks.

Frozen audio encoders let us compare waveform sounds without first dividing them into groups by timing or click count. We identify click and whole-coda forms independently, so their association can be tested instead of assumed. We also use several encoders rather than relying on one model's representation of sound. The direct spectral and envelope tests provide a different kind of check: they ask whether the learned click forms have differences that can be measured in the original recordings. These steps matter because the scientific claim concerns the recorded acoustics, not the clustering algorithm itself.

These forms should nevertheless be understood as acoustic groupings of the available recordings. Their boundaries depend on the representation and grouping resolution, and some of their acoustic properties vary with recording conditions. The results do not establish that whales perceive the same discrete categories. Instead, the forms provide a way to measure acoustic variation and examine how it is organized within and across codas.

\subsection{Different organizing relationships within and across codas}

The key difference between the scales is what the acoustic forms contribute to prediction. Within a coda, knowing one click form helps predict the next, but adding click-form order does not improve the tested classification of the whole coda. The sound and spacing of the clicks remain central to describing that vocalization. Across codas, the question is no longer how a single vocalization is assembled. Instead, the forms of earlier complete codas help predict the form that follows. Both scales contain useful ordering information, but that information answers different acoustic questions.

The sequence result adds a concrete finding to this comparison. Knowing the second preceding whole coda improves prediction beyond knowing only the immediate predecessor. The gain is present across many constructed segments and exceeds two broad first-order references. The stricter segment-local reference does not show an additional excess. Neither does the separate pooled calibration reported in the supplement. These checks do not undo the held-out prediction gain, but they leave open an explanation based on neighboring-coda transitions. The data therefore show a useful relationship across recorded codas without establishing a particular higher-order biological rule.

These findings also show why the two scales should be studied together. A timing-only account misses differences between the sounds of the clicks themselves. A study of isolated codas cannot measure how earlier complete codas help predict later ones. And a single test of order would miss the distinction between predicting the next click and identifying the acoustic form of a whole coda. Examining these questions side by side gives a clearer account of how waveform sound, rhythmic spacing, and sequence order each contribute to the recorded codas.

\section{Limitations}

The analysis uses the CETI recording and annotation ledger, with 1,952 codas from 63 recordings. The stricter click analyses use 1,132 codas after shared-audio and overlapping-window exclusions. The four source pools are conservative exclusion groups, not identified animals or social units. Relative recording times allow constructed sequence segments, but not verified behavioral bout boundaries. Results from these sources do not establish that the same forms and relationships occur in other populations.

The 12 click forms and 32 whole-coda forms are acoustic groupings, not verified biological or perceptual categories. Their boundaries depend on model representations, the chosen resolution, and recording conditions. The row-order correction makes the grouping reproducible under storage permutations, but does not make its categories unique. Date, social-unit, caller, and published coda-type labels are absent from the current ledger, so the earlier biological transfer and type-agreement analyses cannot be retained.

The primary click windows follow annotated positions. Independent detection gives a different event stream and does not reproduce the annotation-based click-order result. Matched-event tests suggest that event inclusion contributes to this discrepancy, but do not independently establish the identities of all detected peaks. The click-order finding should therefore be understood as conditional on the annotation-defined click sample and the available encoder and acoustic controls.

The whole-coda sequence gains are conditional on constructed segments and the recorded data. Source-level uncertainty remains substantial, and the strict local first-order and run-preserving references limit stronger interpretations. Repeated categories, heterogeneous local transitions, and the operational segment boundaries could contribute to the apparent second-predecessor gain. We cannot infer a biological sequence rule or a continuous exchange without additional temporal and behavioral context.

\section{Conclusion}

Click waveforms carry information about sperm-whale codas that click count and timing alone do not describe. The sounds of the clicks and their spacing both matter for distinguishing individual codas. In the annotated sample, neighboring click forms also help predict one another, although adding their order does not improve the tested classification of the complete coda. Across recorded sequences, earlier whole-coda forms help predict which form comes next.

These are different contributions of acoustic information. Within a coda, click waveform composition and rhythmic spacing help describe the sound of a complete vocalization. Across codas, the order of those complete vocalizations provides information about what follows. By measuring click forms and whole-coda forms independently, we can study both relationships without defining either one from timing alone.

\section*{Supplementary material}
The supplementary material gives the full settings, additional checks, and results for click extraction, encoder and clustering choices, direct acoustic tests, sequence controls, detector discrepancies, and reproduction.

\section*{Acknowledgments}
The authors thank the researchers and field teams who collected and annotated the recordings shared through the public collection linked by Gubnitsky and Diamant \citep{gubnitsky2024presence}. We also acknowledge Project CETI and the institutions involved in the broader recording and coda-annotation work. This study is a secondary analysis; the authors did not collect the original audio. This research received no external funding.

\section*{Author declarations}
\subsection*{Conflict of interest}
The authors have no conflicts to disclose.

\subsection*{Ethics approval}
This study is a secondary analysis of previously recorded, publicly available whale sounds and click annotations. The authors conducted no fieldwork or animal interventions.

\subsection*{Use of AI-assisted tools}
The authors used an AI-assisted language tool for editorial restructuring, journal-format conversion, and language refinement. All scientific content, numerical results, citations, interpretations, and final wording were reviewed and verified by the authors, who take full responsibility for the manuscript.

\section*{Data availability}
The source audio and click-time annotations are publicly available in the SpermWhales collection linked in reference 34 of Gubnitsky and Diamant \citep{gubnitsky2024presence} (\url{https://drive.google.com/drive/folders/1HkGZcpYbrft3pGtqdt45bUZXVcrJVrdm}). We analyzed 33 remote WAV recordings and 30 tag FLAC recordings, yielding 1,952 coda crops. The supplementary CSV manifest identifies all 63 recordings and their corresponding annotations by original filename, public file identifier, file size, and SHA-256 checksum. The original recordings remain accessible through the public collection and are not redistributed with this manuscript.

\section*{References}

\begingroup
\setlength{\parskip}{0pt}
\setlength{\topsep}{0pt}
\setlength{\partopsep}{0pt}
\def\theendnotes{}
\bibliography{references}
\endgroup
\end{document}